\documentclass{article}

\usepackage{PRIMEarxiv}

\usepackage[table,xcdraw]{xcolor}
\usepackage[T2A,T1]{fontenc}        
\usepackage[utf8]{inputenc}         
\usepackage[russian,english]{babel} 
\usepackage{libertine}              
\usepackage{microtype}              

\usepackage{amsmath}                
\usepackage{amssymb}                
\usepackage{amsfonts}               

\usepackage{graphicx}               
\graphicspath{{media/}}             
\usepackage{geometry}               
\usepackage{booktabs}               
\usepackage{multirow}               
\usepackage{makecell}               
\usepackage{placeins}               
\usepackage{fancyhdr}               

\usepackage{hyperref}               
\usepackage{url}                    
\usepackage{textcomp}               
\usepackage{gensymb}                
\usepackage{nicefrac}               
\usepackage{lipsum}                 
\usepackage{orcidlink}
\title{Phase-OTDR Event Detection Using Image-Based Data Transformation and Deep Learning}

\author{
  Muhammet Cagri Yeke \orcidlink{0009-0002-6896-1724} \\
  Department of Biotechnology and Bioengineering\\
  Izmir Institute of Technology\\
  Urla, Izmir, Turkey \\
  \texttt{muhammetcagri@iyte.edu.tr} \\
  \And
  Samil Sirin \orcidlink{0000-0002-7106-1519} \\
  Department of Electrical and Electronics Engineering \\
  Izmir Institute of Technology \\
  Urla, Izmir, Turkey \\
  \texttt{samilsirin@iyte.edu.tr} \\
  \And
  Kivilcim Yuksel \orcidlink{0000-0003-1512-3022} \\
  Department of Electrical and Electronics Engineering \\
  Izmir Institute of Technology \\
  Urla, Izmir, Turkey \\
  \texttt{kivilcim.yuksel@iyte.edu.tr} \\
  \And
  Abdurrahman Gumus \orcidlink{0000-0003-2993-5769} \\
  Department of Computer Engineering \\
  Isparta University of Applied Sciences \\
  Isparta, Turkey \\
  \texttt{abdurrahmangumus@isparta.edu.tr} \\
}

\begin{document}
\maketitle

\begin{abstract}
This study focuses on event detection in optical fibers, specifically classifying six events using the Phase-OTDR system. A novel approach is introduced to enhance Phase-OTDR data analysis by transforming 1D data into grayscale images through techniques such as Gramian Angular Difference Field, Gramian Angular Summation Field, and Recurrence Plot. These grayscale images are combined into a multi-channel RGB representation, enabling more robust and adaptable analysis using transfer learning models. The proposed methodology achieves high classification accuracies of 98.84\% and 98.24\% with the EfficientNetB0 and DenseNet121 models, respectively. A 5-fold cross-validation process confirms the reliability of these models, with test accuracy rates of 99.07\% and 98.68\%. Using a publicly available Phase-OTDR dataset, the study demonstrates an efficient approach to understanding optical fiber events while reducing dataset size and improving analysis efficiency. The results highlight the transformative potential of image-based analysis in interpreting complex fiber optic sensing data, offering significant advancements in the accuracy and reliability of fiber optic monitoring systems. The codes and the corresponding image-based dataset are made publicly available on GitHub to support further research: \url{https://github.com/miralab-ai/Phase-OTDR-event-detection}.

\end{abstract}


\section{Introduction}

Event detection and classification involve identifying, analyzing, and categorizing patterns within event streams to provide insights into real-world occurrences.\cite{LIU2021104380,porumb2020precision} This process finds applications across diverse fields, from emergency response and crisis management to monitoring trends and potential threats on social media platforms.\cite{KIM2020107092, AHMAD2022116626, s22124531, KIM2022102789}

Modern sensor systems enhanced by deep learning techniques serve a wide spectrum of event detection applications. Deep learning models facilitate precise analysis in various domains: acoustic sensors enable noise source identification in urban environments,\cite{s21227470, ijerph20043683} image processing systems detect object movements and monitor traffic flow,\cite{ionescu2019object, 9844855, math10060873} seismic sensors classify earthquakes based on magnitude and location,\cite{9112316, Lomax2019} and wearable devices detect medical events such as heart arrhythmias or epileptic seizures.\cite{s22051776, Tang2021, Beniczky2021}

Phase-sensitive Optical Time-Domain Reflectometer (Phase-OTDR) represents a distributed sensing technology that monitors acoustic events along optical fiber. Unlike conventional OTDR systems, Phase-OTDR utilizes coherent light sources to measure phase information of backscattered lightwaves. Vibrations or disturbances cause regional changes in fiber length and refractive index, enabling spatially resolved event localization and characterization. This distributed sensing capability makes Phase-OTDR particularly valuable for real-time monitoring of large areas including pipelines,\cite{pipe} railways,\cite{rail} and security systems.\cite{sec}

The ability to track phase changes along an optical fiber as a function of position and time offers a wide range of event detection possibilities for Phase-OTDR. However, interpreting these phase traces is challenging due to inherent noise sources and the similar strain characteristics exhibited by the fiber for different events. Consequently, significant effort has been dedicated to accurately recognizing event types from phase traces. Historically, signal processing techniques such as wavelet transform,\cite{wavelet} Fourier analysis,\cite{jason2017laboratory} and statistical methods\cite{stat} were instrumental in this recognition process. Nevertheless, the field has evolved, with machine learning methods gaining prominence. These data-driven approaches excel at discerning complex patterns, leading to their increasing dominance and marking a significant shift in event detection methodologies.

\subsection{Related Works}

Machine learning and deep learning have significantly advanced Phase-OTDR event detection and classification, with numerous studies exploring various models and preprocessing techniques. For instance, Cao et al. \cite{CAO2023100372} utilized SVM and CNN with spatio-temporal matrix normalization, achieving 94.0\% accuracy. Chen et al. \cite{electronics12183757} employed a Dendrite Net (DD) with VTTCCG, reaching 98.6\% accuracy after normalization and spatio-temporal feature extraction. Gan et al. \cite{Gan22024} applied transfer learning with VGGish and SVM, while Hu et al. \cite{HU2024130818} used AlexNet within a TSC framework, incorporating normalization and StyleGAN for data augmentation. Kamanga et al. \cite{KAMANGA2024104032} achieved 98.2\% accuracy with a modified AlexNet using MFCC-DP and active learning. Li et al. \cite{Li2024Aaa} introduced a semi-supervised MT-ACNN-SA-BiLSTM model. Additionally, Sasi et al. \cite{10627811} used LightGBM with feature selection, and Wang et al. \cite{10557760} explored traditional machine learning models like SVM, RF, KNN, and NB. Recent advancements in 2025 include Duan et al.'s \cite{DUAN2025104171} ISAT model, which uses a 1D CNN and a modified Transformer Encoder for high accuracy. Hu et al. \cite{HU2025131393} proposed SDENet with Gaussian noise addition and spatiotemporal maps. Li et al. \cite{10877714} introduced DASFormer, a Transformer-based model for long sequence classification, achieving 99.6\% accuracy. Luo et al. \cite{10876181} utilized knowledge distillation with a 4-layer CNN and ResNet-34. Wang et al. \cite{11128983} developed an IRMS-CNN with Contrastive Prototype Learning, and Cheng et al. \cite{Chengaa2025} presented $\Phi$-GLMAE, a Global-Local Masked Autoencoder. A comprehensive overview of these studies, including their preprocessing methods, architectures, dataset configurations, and reported accuracies, is provided in Table \ref{tab:dl_models_full}.

\subsection{Literature Gaps}

Despite significant progress in Phase-OTDR event classification, critical gaps persist. While some recent studies explore image-based transformations of 1D time series data, there is a notable lack of research focusing on combining multiple distinct image transformation techniques to create rich, multi-channel image representations. Existing approaches often rely on single transformation methods or basic deep learning architectures, overlooking the potential for integrating diverse feature perspectives. Furthermore, a comprehensive comparative analysis of how different image transformation methods, especially when combined, impact classification performance within the Phase-OTDR domain remains underexplored. Crucially, the potential for these advanced image-based transformations to enhance data management efficiency by significantly reducing the raw dataset size has not been thoroughly investigated as a primary benefit. Addressing these gaps is essential for developing more robust, efficient, and highly accurate Phase-OTDR event detection systems.

\subsection{Motivation and Study Outline}

\begin{figure}[!ht]
\centering
\includegraphics[width=0.95\linewidth]{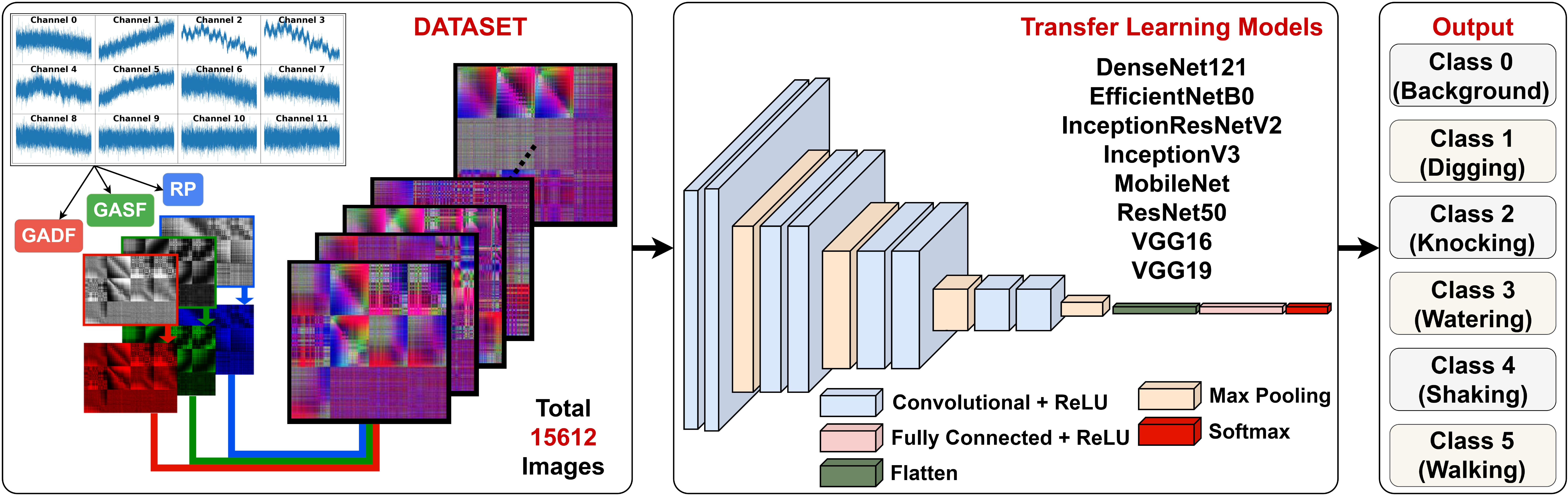}
\caption{Transforming 1D Phase-OTDR data into images and employing deep learning models for event classification based on created dataset.}
\label{fig:framework}
\end{figure}

This study addresses the identified literature gaps by introducing a novel image-based transformation approach for enhanced Phase-OTDR data analysis. Our motivation stems from the need to overcome the limitations of traditional 1D signal processing and to leverage the capabilities of powerful pre-trained deep learning models fully. We propose a methodology that uniquely employs Gramian Angular Difference Field (GADF), Gramian Angular Summation Field (GASF), and Recurrence Plots (RP) to create multi-channel RGB image representations from 1D time series data. This innovative transformation facilitates advanced feature extraction, improved event discrimination capabilities, and offers the potential for more efficient data management. Figure \ref{fig:framework} illustrates the overall framework of transforming 1D Phase-OTDR data into images and employing deep learning models for event classification.

\subsection{Novelties and Contributions}

This study presents a framework for event detection in optical fibers, addressing the limitations of traditional Phase-OTDR data analysis. Our primary contributions are multifaceted:

\begin{itemize}
    \item We introduce a methodology for converting one-dimensional Phase-OTDR data into two-dimensional grayscale images, forming the basis for an image-based analysis approach. This transformation enhances the visual representation of events, facilitating more effective feature extraction and pattern recognition.
    \item A color allocation process is developed to integrate multiple grayscale images into a single multi-channel RGB representation. This approach enriches the data's information density, providing a more comprehensive input for deep learning models.
    \item The proposed image transformation and integration significantly reduce the dataset size while preserving critical information. This contributes to more efficient data handling and processing, which is crucial for real-time applications.
    \item The research explores the efficacy of transfer learning for Phase-OTDR event classification using established CNN architectures, including 8 transfer learning models. This involves a comparative analysis of these different models, alongside an evaluation of fine-tuning versus feature extraction strategies, demonstrating a robust and adaptable approach.
\end{itemize}

This paper presents a novel image-based analysis approach for Phase-OTDR event classification using multi-channel data transformation and deep learning. The remainder is organized as follows: Section \ref{sec:2} describes the methodology including data collection, dataset details, image transformation techniques, and deep learning models. Section \ref{sec:3} presents results and performance analysis. Section \ref{sec:4} concludes the study.

\section{Methods}\label{sec:2} 

This study presents a comprehensive methodology for Phase-OTDR event classification using image-based data transformation and transfer learning approaches. The methodology utilizes the publicly available dataset from Cao et al.\cite{CAO2023100372} and applies novel image transformation techniques to convert 1D time series data into multi-channel representations suitable for deep learning analysis.

\subsection{Data Collection Setup}\label{sec:2.1}

\begin{figure}[!ht]
    \centering
    \includegraphics[width=0.8\linewidth]{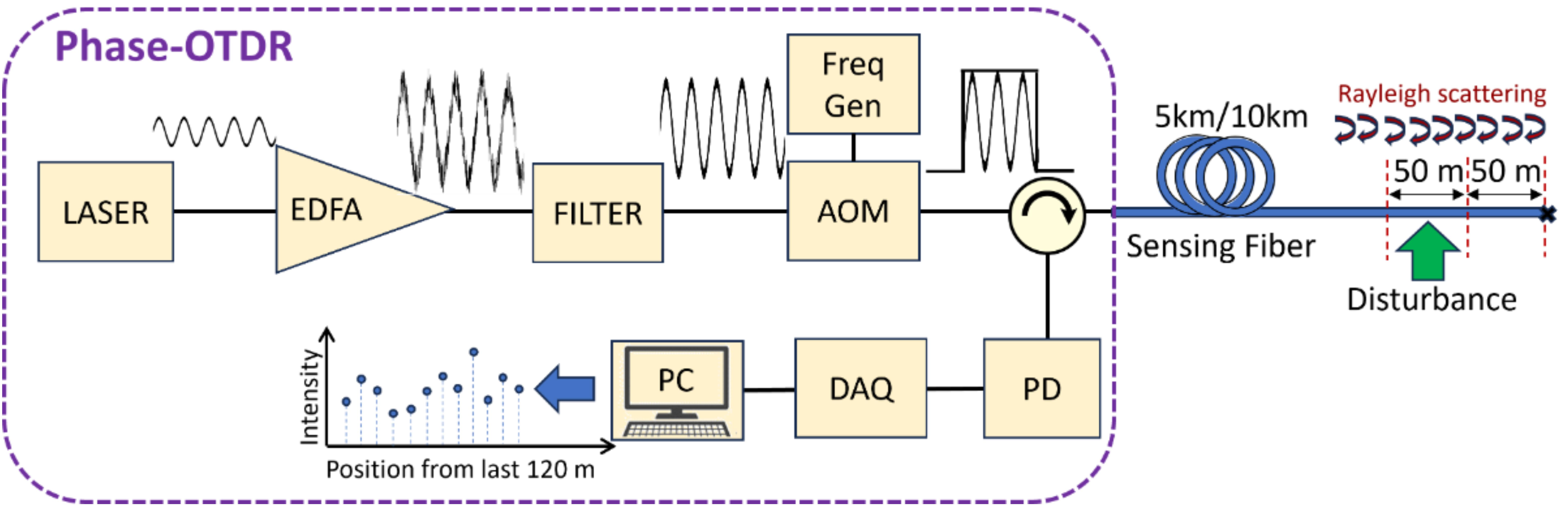}
    \caption{The setup of the Phase-OTDR system for event detection (adapted from Cao et al.).}
    \label{fig:setup}
\end{figure}

This study utilizes the Phase-OTDR dataset collected by Cao et al.\cite{CAO2023100372} using their experimental setup. The original measurement system, as described by Cao et al., is illustrated in Figure \ref{fig:setup}. Light generated by the laser is amplified in an Erbium-Doped Fiber Amplifier (EDFA) and filtered to eliminate amplified spontaneous emission (ASE) noise. An Acousto Optic Modulator (AOM) driven by a function generator converts continuous light to 400 ns pulses with 40 m spatial resolution and 12.5 or 8 kHz repetition frequency. The sensing fiber comprises three sections: a 5-10 km leading fiber kept isolated from vibration in a soundproof box, and two 50 m armored sections where the second section is exposed to vibrational events. Rayleigh backscattered light is guided by the circulator to the photodetector for intensity detection and data acquisition.

\FloatBarrier

\subsection{Dataset Description and Preprocessing}\label{sec:2.2}

\begin{figure}[!ht]
    \centering
    \includegraphics[width=0.95\linewidth]{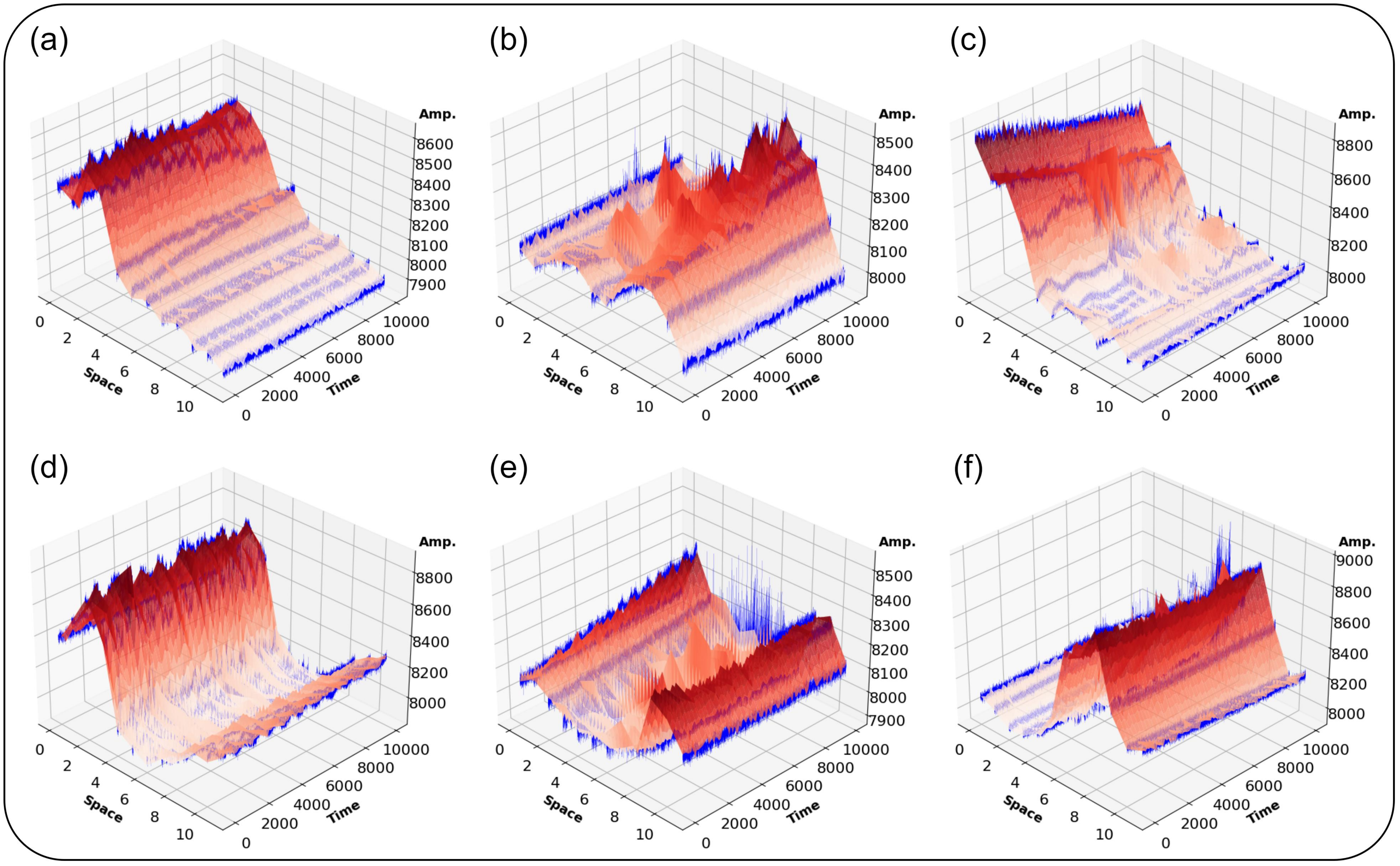}
    \caption{Interpolated spatial-temporal samples showing distinctive characteristics of different events: (a) Background, (b) Digging, (c) Knocking, (d) Watering, (e) Shaking, and (f) Walking. Raw data without preprocessing demonstrates natural event characteristics with blue lines indicating 12 distinct fiber regions.}
    \label{fig:data3D}
\end{figure}

The dataset comprises 15,612 samples collected by sending successive 10,000 pulses and detecting backscattered light intensity from the last 120 m fiber section.\cite{CAO2023100372} Each measurement contains 12 intensity values of Rayleigh backscattered signal from 12 equidistant spatial points on the 120 m fiber section, forming a 12×10,000 intensity matrix per sample. During each measurement, one of six distinct disturbance events was applied on the second fiber section: background, digging, knocking, watering, shaking, and walking. Figure \ref{fig:data3D} presents sample measurement data for each event type.

\FloatBarrier

The dataset was originally divided into training and test sets with an 8:2 ratio. Event measurements are stored in .mat format files corresponding to each event type. Table \ref{tab:t1} presents the distribution of samples across different event categories, showing the sample counts and corresponding labels used for classification.

\begin{table}[!ht]
\centering
\caption{Distribution of dataset samples across different event types with corresponding labels.}
\begin{tabular}{lll}
\hline
\label{tab:t1}
\textbf{Event Categories} & \textbf{Sample Counts} & \textbf{Labels} \\ \hline
Background                  & 3094                   & 0               \\
Digging                     & 2512                   & 1               \\
Knocking                    & 2530                   & 2               \\
Watering                    & 2298                   & 3               \\
Shaking                     & 2728                   & 4               \\
Walking                     & 2450                   & 5               \\ \hline
Total                       & 15612                  & 6               \\ \hline
\end{tabular}
\end{table}
\FloatBarrier

\subsection{Image-Based Data Transformation Techniques for Analyzing 1D Phase-OTDR Data}\label{sec:2.3}

Phase-OTDR data is represented as 1D time series comprising amplitude information acquired within specified time intervals. Converting this 1D time series data into image representations enables utilization of advanced computer vision techniques and pre-trained deep learning models.\cite{Gabriel22, Woodward23} Three mathematical transformation techniques are employed: Gramian Angular Difference Field (GADF), Gramian Angular Summation Field (GASF), and Recurrence Plot (RP).\cite{batista2023intelligent, s23208592, 9486862, 10185559}

This approach transforms 1D Phase-OTDR data into visually interpretable images using mathematical transformations, including GADF, GASF, and RP. This conversion simplifies the analysis of intricate 1D data by providing a visual representation of monitored events. Furthermore, by employing these three distinct methods, a multi-channel RGB image is constructed, where each channel corresponds to a different transformation. This multi-channel representation facilitates a comprehensive analysis by integrating diverse feature perspectives. Meaningful features are then extracted from these RGB image data using transfer learning model architectures, which enhances the classification accuracy of fiber optic sensor data. This image-based analysis is a key aspect of the methodology, designed to improve Phase-OTDR data analysis capabilities.

Each technique captures distinct temporal characteristics: GADF emphasizes temporal variations and differences between data points, GASF highlights cumulative patterns and temporal correlations, while RP reveals recurrent behaviors and periodic structures in the time series. These complementary representations provide comprehensive feature extraction from different mathematical perspectives.

\begin{figure}[!ht]
    \centering
    \includegraphics[width=0.95\linewidth]{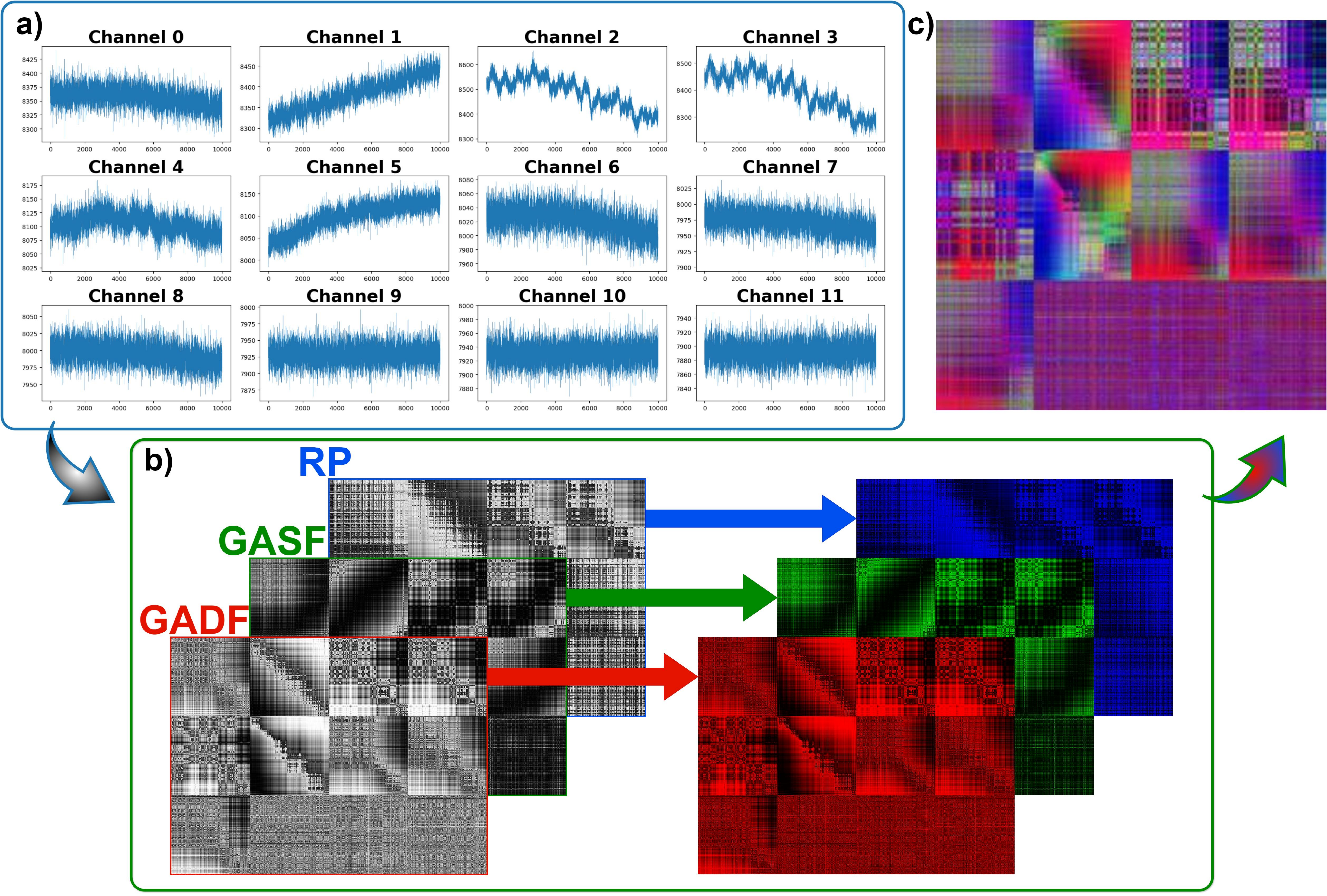}
    \caption{Image-based data transformation workflow for Phase-OTDR event detection. (a) Raw 1D time series data from 12 spatial fiber regions, (b) Transformation into individual 500×500 pixel images using GADF, GASF, and RP techniques organized in 3×4 grid, (c) Final RGB images with downsampled 224×224 resolution for deep learning compatibility.}
    \label{fig:imagetransform}
\end{figure}
\FloatBarrier

The transformation process converts data from 12 spatial points into individual 500×500 pixel images using each technique. These images are organized into a 3×4 grid format, creating 1500×2000 pixel grayscale representations. Each technique's output is allocated to specific RGB channels (GADF→Red, GASF→Green, RP→Blue), producing multi-channel color images. Final images are downsampled to 224×224 pixels for compatibility with pre-trained deep learning architectures, as illustrated in Figure \ref{fig:imagetransform}.

\subsubsection{Gramian Angular Field}\label{sec:2.3.1}

The Gramian Angular Field (GAF) is a powerful technique for analyzing time series data by converting it into a visual format that facilitates the exploration of temporal relationships.\cite{2015arXiv150907481W} This transformation involves representing the data in polar coordinates, where angular information encodes the relationships between data points, and radial values represent temporal data. Three key components of GAF are the Gram Matrix, GASF Matrix, and GADF Matrix. 

The Gram Matrix, also known as the Gramian Matrix, is a fundamental mathematical construct in GAF. It is a matrix representation of the inner products of the data vectors. If we have a dataset with vectors, $x_1, x_2, \ldots, x_n$, the Gram Matrix $G$ is constructed as follows:

\begin{align}
G &= X^T X \\
G &= \begin{bmatrix}
(x_1,x_1) & (x_1,x_2) & \cdots & (x_1, x_n) \\
(x_2,x_1) & (x_2,x_2) & \cdots & (x_2, x_n) \\
\vdots & \vdots & \ddots & \vdots\\
(x_n,x_1) & (x_n,x_2) & \cdots & (x_n, x_n)
\end{bmatrix}
\end{align}

Here, $X^T$ represents the transpose of the data matrix $X$. The Gram Matrix $G$ summarizes the pairwise inner products between data points and serves as a crucial step in the GAF transformation.

GASF Matrix is derived from the Gram Matrix. It captures the cosine of the summation of angular values between data points. Each element of the GASF Matrix corresponds to a pair of data points and is calculated as:

\begin{align}
\text{GASF}(x_i,x_j) &= \cos(\Theta_i + \Theta_j) \\
\text{GASF} &= \begin{bmatrix}
\cos(\Theta_1 + \Theta_1) & \cos(\Theta_1 + \Theta_2) & \cdots & \cos(\Theta_1 + \Theta_j) \\
\cos(\Theta_2 + \Theta_1) & \cos(\Theta_2 + \Theta_2) & \cdots & \cos(\Theta_2 + \Theta_j) \\
\vdots & \vdots & \ddots & \vdots \\
\cos(\Theta_i + \Theta_1) & \cos(\Theta_i + \Theta_2) & \cdots & \cos(\Theta_i + \Theta_j)
\end{bmatrix}
\end{align}

where $\Theta_i$ and $\Theta_j$ are the angular values associated with two specific data points. The GASF Matrix emphasizes cumulative patterns and relationships between data points, providing insights into how they collectively evolve over time.

The GADF Matrix is another component of the GAF transformation. It is also derived from the Gram Matrix and calculates the cosine of the difference between angular values of data points. Each element of the GADF Matrix is calculated as:

\begin{align}
\text{GADF}(x_i,x_j) &= \sin(\Theta_i - \Theta_j) \\
\text{GADF} &= \begin{bmatrix}
\sin(\Theta_1 - \Theta_1) & \sin(\Theta_1 - \Theta_2) & \cdots & \sin(\Theta_1 - \Theta_j) \\
\sin(\Theta_2 - \Theta_1) & \sin(\Theta_2 - \Theta_2) & \cdots & \sin(\Theta_2 - \Theta_j) \\
\vdots & \vdots & \ddots & \vdots \\
\sin(\Theta_i - \Theta_1) & \sin(\Theta_i - \Theta_2) & \cdots & \sin(\Theta_i - \Theta_j)
\end{bmatrix}
\end{align}

\begin{figure}[!ht]
    \centering
    \includegraphics[width=0.95\linewidth]{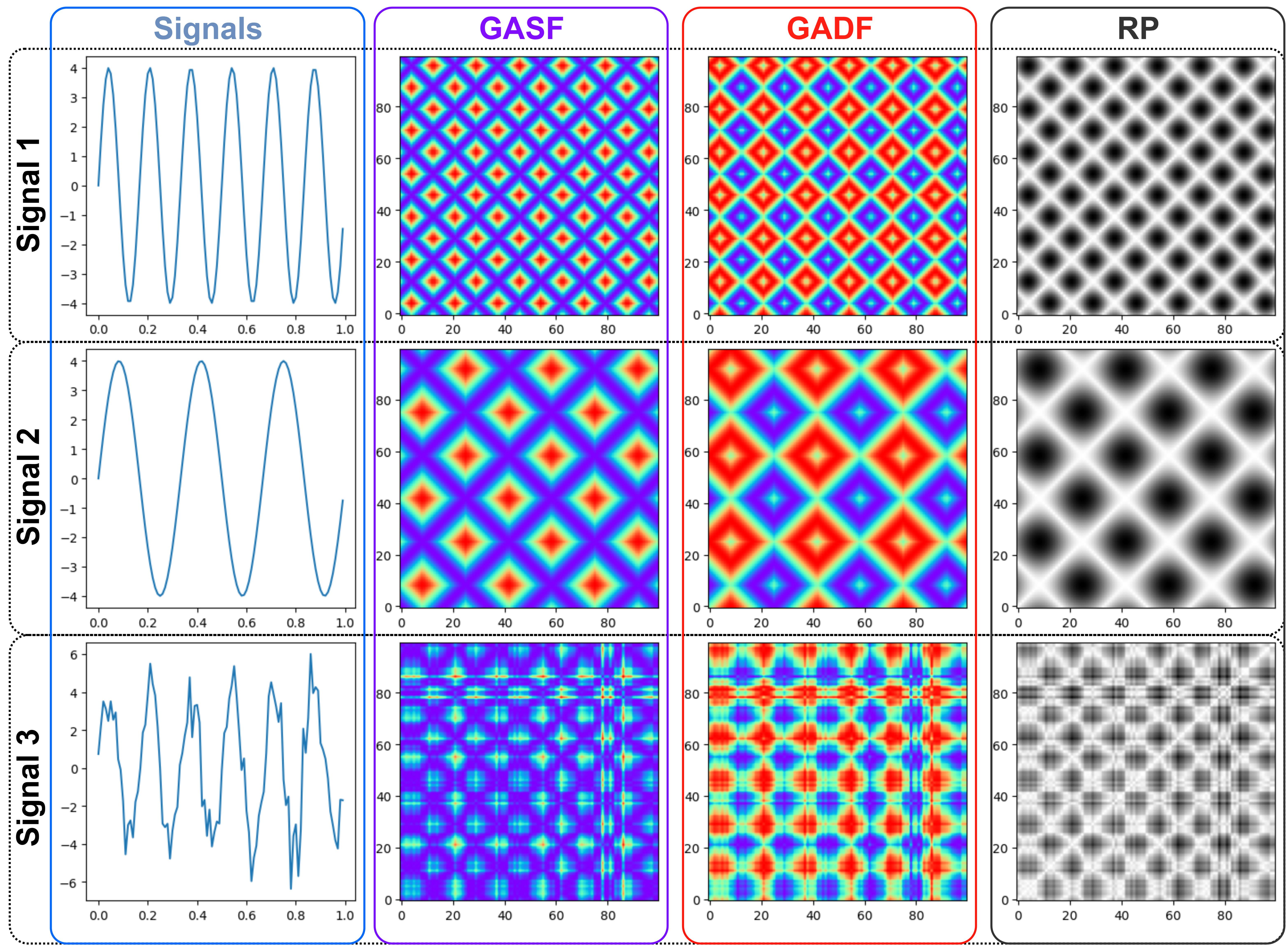}
    \caption{Exploring signal transformations: GASF, GADF, and RP image encoding methods unveil varied behaviors across different sinusoidal signals. Signal 1: A pure sinusoidal signal with an amplitude of 4 and a frequency of 6 Hz. Signal 2: Another sinusoidal signal with an amplitude of 4 and a frequency of 3 Hz. Signal 3: Similar to Signal 1 but with added random noise, creating a sinusoidal signal with an amplitude of 4 and a frequency of 6 Hz.}
    \label{fig:image_sinusoidal_signals}
\end{figure}

The GADF Matrix highlights variations and deviations in temporal relationships between data points, shedding light on how data points differ from each other. Visual examples of GASF and GADF transformations for different sinusoidal signals are presented in Figure \ref{fig:image_sinusoidal_signals}.

\FloatBarrier

\subsubsection{Recurrence Plot}\label{sec:2.3.2}

Recurrence Plot (RP) introduced by Eckmann and his colleagues in 1987 for visualizing and analyzing recurrent behaviors and structures in dynamic systems and time series data.\cite{Eckmann_1987} It has found widespread use, particularly in understanding complex systems and discovering specific behaviors. The fundamental mathematical basis of the RP involves measuring the similarity or proximity between two time points. If the distance between two points falls below a specified threshold value, the two points are considered close to each other. Mathematically, the distance $d(i, j)$ between two time points $i$ and $j$ is calculated as follows:
 
\begin{align}
d(i, j) &= ||x(i) - x(j)||
\end{align}

Here, $x(i)$ and $x(j)$ represent the values of time series data at indices $i$ and $j$, respectively. 

This process is repeated for all data points in the time series, resulting in a matrix. This matrix forms the basis of the RP. The $i$-th row and $j$-th column of the matrix represent the distance between time points $i$ and $j$. If this distance is less than a specified threshold ($\epsilon$), the corresponding element in the matrix is set to 1; otherwise, it is set to 0. 
The mathematical representation of the RP can be expressed as:

\begin{align}
R_{i, j} &= 
\begin{cases} 
1 & \text{if } d(i, j) < \epsilon \\
0 & \text{otherwise}
\end{cases}
\end{align}

Here, $R_{i, j}$ is an element of the matrix that indicates the similarity between the $i$-th and $j$-th time points. $\epsilon$ serves as a threshold, determining how close two points in time need to be. This matrix constitutes the RP.

\subsection{Convolutional Neural Networks (CNNs)}\label{sec:2.4}

CNNs are computational models designed for learning from image-based data.\cite{9451544, 2015arXiv151108458O} These networks comprise two stages: feature extraction and classification. Convolutional and pooling layers handle feature extraction. Convolutional layers perform convolution operations using filters to process input data:

\begin{align}
x_{n, i} &= f \left( \sum_{j=1}^{N} W_{n-1, k} * y_{n-1, m} + b_{n, k} \right) \\
\text{MaxPooling}(x) &= \max_{i,j \in R} x_{i,j} \\
\text{ReLU}(x) &= \begin{cases} 0 & \text{if } x < 0 \\ x & \text{if } x \geq 0 \end{cases} \\
S(y_i) &= \frac{e^{y_i}}{\sum_{j=1}^{K} e^{y_j}}
\end{align}

where $x_{n, i}$ represents the $i$th feature map, $W_{n-1, k}$ and $b_{n, k}$ are weights and biases of the $k$th filter, and $y_{n-1, m}$ is the $m$th feature map from the preceding layer. Pooling layers reduce feature map dimensions while preserving essential information. Fully connected layers use ReLU activation for feature processing and softmax for multi-class classification.\cite{8016501}

\subsubsection{Transfer Learning and Model Validation}\label{sec:2.4.1}

Transfer learning transfers knowledge acquired in one domain to other domains, reducing resource-intensive requirements such as large datasets and high computational power typically needed for training deep learning models.\cite{9134370, ZHAO2024122807} It has wide applications across computer vision, natural language processing, and healthcare.\cite{Wang2023, taiar2022fine, 9076082} All pre-trained models are initially loaded with ImageNet weights, and their final classification layers are replaced with 6 neurons corresponding to Phase-OTDR event categories.

Two implementation strategies are evaluated for layer configuration. When setting layers as "Trainable: True," these layers retain their initial weights from pre-training but are fine-tuned using the new dataset. This increases the number of trainable parameters, leading to longer learning process and higher computational load. It is advantageous when the new dataset is relatively small and closely related to the original pre-training dataset. Conversely, selecting "Trainable: False" freezes these layers as feature extractors, preventing weight updates during training. This significantly reduces trainable parameters and expedites training but potentially constrains adaptability. This option is beneficial when the new dataset is significantly different or when the original dataset is vast compared to the new one.

The transfer learning architectures evaluated in this study are summarized in Table \ref{tab:models}.

\begin{table}[ht]
\caption{Summary of deep learning models, including their architectures, advantages, disadvantages, and applications. Abbreviations: VGG - Visual Geometry Group, ResNet - Residual Network, Inception - Inception Network, IncResNetV2 - Inception Residual Network V2, MobileNet - Mobile Neural Network, EfficientNet - Efficient Neural Network, DenseNet - Densely Connected Convolutional Network. Each model is characterized by its unique design and practical use cases across various domains.}
\label{tab:models}
\resizebox{\columnwidth}{!}{%
\begin{tabular}{l|l|l|l}
\hline
\multicolumn{1}{c|}{\textbf{Model}} &
  \multicolumn{1}{c|}{\textbf{Architecture}} &
  \multicolumn{1}{c|}{\textbf{Advantages \& Disadvantages}} &
  \multicolumn{1}{c}{\textbf{Applications}} \\ \hline
VGG-16 &
  \makecell[l]{16-layer deep convolutional network with \\ fixed 3x3 filters and three fully connected \\ layers. Approximately 138M parameters \cite{simonyan2014very}.} &
  \makecell[l]{Simple and effective, but requires high \\ memory and computational power.} &
  \makecell[l]{Fall Detection \cite{Chhetri2021}. \\ POF Weight Detection \cite{10189408}. \\ Optical Measurement Technology \cite{Lu_2025_msat}.} \\ \hline
VGG-19 &
  \makecell[l]{Deeper version of VGG-16 with 19 layers \\ and approximately 144M parameters \\ \cite{simonyan2014very}.} &
  \makecell[l]{Higher accuracy but more computationally \\ expensive.} &
  \makecell[l]{FBG Tactile Recognition \cite{LYU2023112906}. \\ Fiber Performance Prediction \cite{10767412}. \\ Image-based Classification \cite{Abbas2023}.} \\ \hline
ResNet50 &
  \makecell[l]{50-layer network with residual blocks and skip \\ connections to address vanishing gradient \\ problem. About 25.6 million parameters \cite{he2016deep}.} &
  \makecell[l]{Enables very deep networks but can be \\ overkill for small datasets.} &
  \makecell[l]{Broken Wire Identification \cite{Wang2024PoolConv}. \\ Weld Defect Recognition \cite{Zhao2024A}. \\ Vibration Recognition \cite{YAN2025112083}.} \\ \hline
InceptionV3 &
  \makecell[l]{Uses parallel convolutional layers with \\ varying kernel sizes for feature extraction. \\ Approximately 23.5M parameters \cite{szegedy2016rethinking}.} &
  \makecell[l]{Efficient and accurate, but complex to \\ implement.} &
  \makecell[l]{Perimeter Security Detection \cite{LYU2021106377}. \\ Pathology Classification \cite{Lilili22}. \\ Fiber Security Detection \cite{WANG2025102802}.} \\ \hline
IncResNetV2 &
  \makecell[l]{Combines inception modules with residual \\ connections. Approximately 55.8M parameters \\ \cite{szegedy2017inception}.} &
  \makecell[l]{High accuracy but computationally \\ expensive.} &
  \makecell[l]{Photonic DNN Acceleration \cite{app11136232}. \\ CLS Disease Detection \cite{s24165398}. \\ Modulation Format Recognition \cite{s24227291}.} \\ \hline
MobileNetV2 &
  \makecell[l]{Lightweight architecture with depthwise \\ separable convolutions. Approximately 3.4M \\ parameters \cite{sandler2018mobilenetv2}.} &
  \makecell[l]{Efficient for mobile and embedded \\ devices, but less accurate for complex tasks.} &
  \makecell[l]{Feature Extraction \cite{Mayrose22}. \\ POF Motion Recognition \cite{10130078}. \\ Fiber Defect Detection \cite{10609799}.} \\ \hline
EfficientNet-B0 &
  \makecell[l]{Balances network depth, width, and \\ resolution using compound scaling. \\ Approximately 5.3M parameters \cite{tan2019efficientnet}.} &
  \makecell[l]{Highly efficient but requires careful \\ tuning.} &
  \makecell[l]{Wavefront Detection \cite{Li22222}. \\ Deep Learning Regression \cite{Igel2023}. \\ Wavefront Sensing \cite{s25020480}.} \\ \hline
DenseNet121 &
  \makecell[l]{Connects each layer to every other layer \\ to improve feature reuse. Approximately \\ 8M parameters \cite{huang2017densely}.} &
  \makecell[l]{Efficient and reduces overfitting, but \\ memory intensive.} &
  \makecell[l]{Vibration Pattern Recognition \cite{PAN2022168127}. \\ DAS Event Recognition \cite{Kayan2023}. \\ Intrusion Signal Recognition \cite{SHEN2025104227}.} \\ \hline
\end{tabular}%
}
\end{table}
\FloatBarrier

\begin{figure}[!ht]
    \centering
    \includegraphics[width = 1\linewidth]{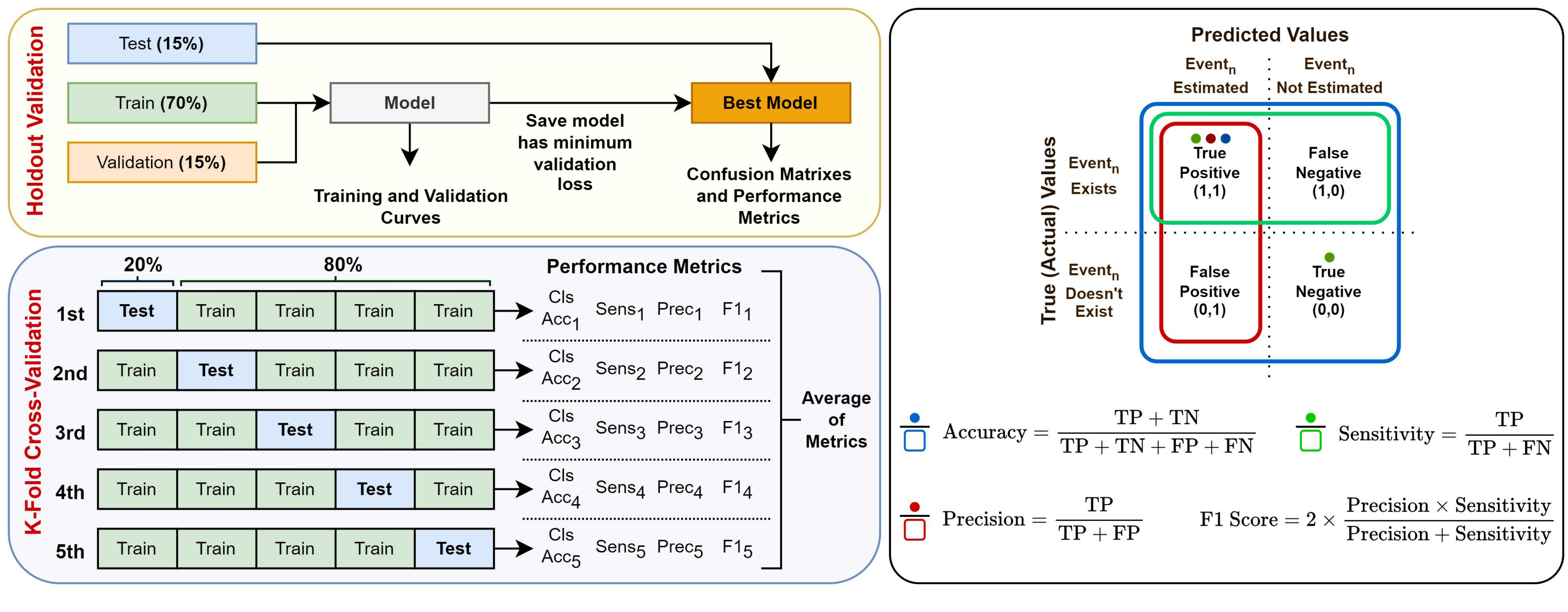}
    \caption{Overview of model validation techniques and performance metrics. The left panel illustrates holdout validation and 5-fold cross-validation approaches, while the right panel shows the calculation of performance metrics, including accuracy, sensitivity, precision, and F1 score, for classification evaluation.}
    \label{fig:combined_validation_metrics}
\end{figure}

Model validation was performed using two techniques: holdout validation and 5-fold cross-validation. Holdout validation divided the dataset into training (70\%), test (15\%), and validation (15\%) subsets. In contrast, 5-fold cross-validation split the data into five folds, with each fold serving as the test set once while the remaining four folds were used for training. This approach provided a more robust evaluation by averaging performance metrics across all folds. Performance metrics were selected to address the challenges of multi-class classification with potential class imbalances. Accuracy measured overall correctness, while sensitivity (recall) assessed the true positive detection rate. Precision quantified the reliability of positive predictions, and the F1 Score provided a harmonic mean of precision and recall. These validation techniques and performance metrics are illustrated in Figure~\ref{fig:combined_validation_metrics}, which provides insights into their application and importance for Phase-OTDR event classification.

\FloatBarrier

\section{Results and Discussion}\label{sec:3}

This study investigates the effectiveness of an image-based deep learning approach for event detection in Phase-OTDR data. The inherent challenges in interpreting phase traces due to noise and similar strain characteristics for different events have motivated a shift towards image-based deep learning. Our approach transforms 1D Phase-OTDR data into multi-channel RGB images using mathematical transformations such as GADF, GASF, and RP, enabling the extraction of meaningful features via transfer learning models. This innovative methodology aims to significantly advance Phase-OTDR data analysis.

\begin{figure}[!ht]
    \centering
    \includegraphics[width=0.8\linewidth]{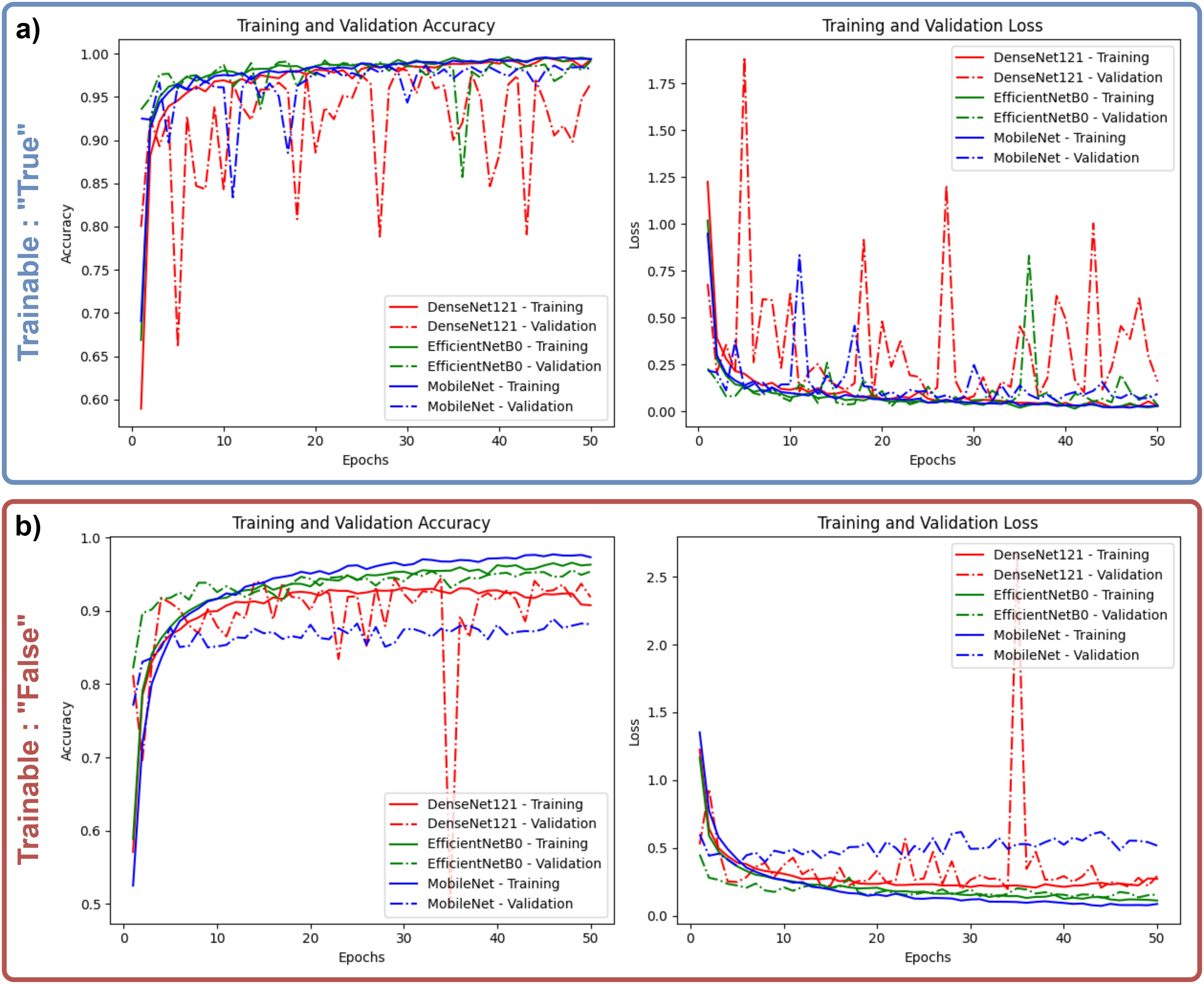}
    \caption{Illustrates the training and validation outcomes of three transfer learning models: DenseNet121, EfficientNetB0, and MobileNet. a) Shows the training and validation results for the transfer learning fine-tuning model where trainable layers were set to "True" for each model. b) Represents the results for the transfer learning model used as a feature extractor with trainable layers set to "False". These visualizations provide a comparative analysis of the models' behaviors under different configurations of trainable layers.}
    \label{fig:train-res}
\end{figure}

The training and validation outcomes for three transfer learning models—DenseNet121, EfficientNetB0, and MobileNet—are presented in Figure \ref{fig:train-res}. These results were obtained using holdout validation under two distinct configurations: with trainable layers set to "True" (fine-tuning) and "False" (feature extraction). As depicted in Figure \ref{fig:train-res}.a, when trainable layers were enabled, all models demonstrated rapid learning and achieved high peak accuracies. For instance, DenseNet121's training accuracy rapidly increased from an initial 58.94\% to a peak of 99.54\%, with validation accuracy similarly rising from 79.96\% to 98.72\%. EfficientNetB0 and MobileNet exhibited comparable performance trajectories, reaching peak training accuracies of 99.61\% and 99.58\%, and validation accuracies of 99.31\% and 98.68\%, respectively. The swift convergence from relatively lower initial accuracies to near-perfect peak values highlights the effectiveness of fine-tuning pre-trained models on our image-transformed Phase-OTDR dataset. This rapid improvement indicates that the models quickly adapted their pre-learned features to the specific patterns of our event data, demonstrating strong learning capabilities. Conversely, Figure \ref{fig:train-res}.b illustrates the performance when trainable layers were frozen. While still achieving respectable accuracies, a noticeable reduction in peak performance and a slower learning trajectory were observed across all models. DenseNet121, for example, reached a peak training accuracy of 93.16\% and validation accuracy of 94.48\%. EfficientNetB0 achieved 96.58\% training and 95.59\% validation accuracy, while MobileNet reached 97.71\% training and 88.91\% validation accuracy. The difference in performance between the "True" and "False" configurations underscores the critical role of fine-tuning in adapting generic pre-trained features to the nuanced characteristics of Phase-OTDR event images, leading to superior classification performance.

\FloatBarrier

\begin{figure}[!ht]
    \centering
    \includegraphics[width=0.95\linewidth]{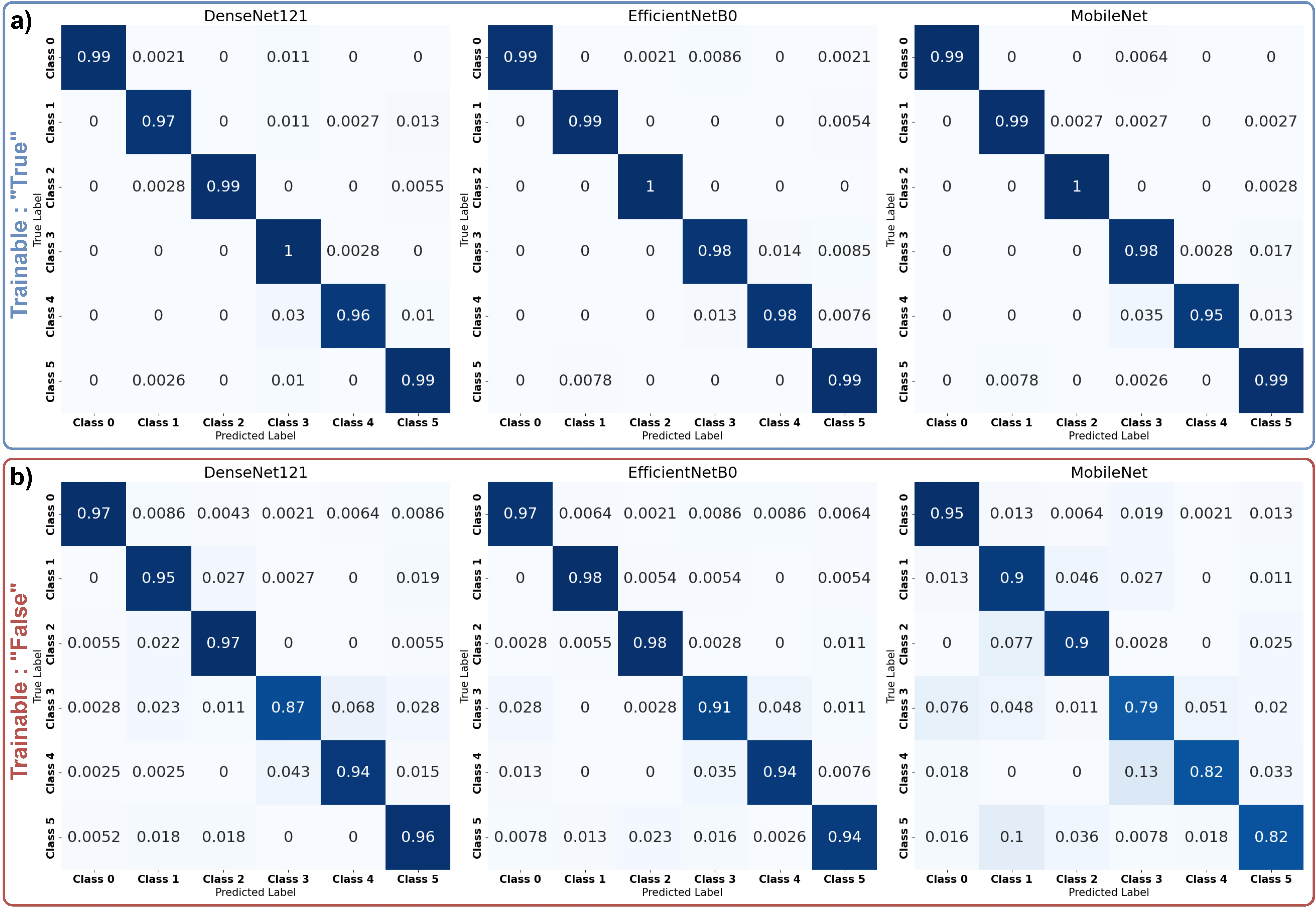}
    \caption{Illustrates the classification performance of the selected DenseNet121, EfficientNetB0, and MobileNet models under "Trainable: True" and "Trainable: False" settings.}
    \label{fig:conf-mat}
\end{figure}

Figure \ref{fig:conf-mat} presents the confusion matrices for DenseNet121, EfficientNetB0, and MobileNet, illustrating their classification performance on the test dataset under holdout validation. A clear visual distinction is observed, with models operating in the fine-tuning mode ("Trainable: True") demonstrating superior event classification compared to the feature extraction mode ("Trainable: False"). This indicates that allowing the models to adapt their pre-trained weights to the specific characteristics of Phase-OTDR event images significantly improves their ability to accurately distinguish between different events. The enhanced performance in fine-tuning mode is crucial for real-world Phase-OTDR applications, as it translates directly to more reliable detection of critical events like digging or walking, minimizing false alarms and missed detections.

\FloatBarrier

Table \ref{tab:t2} provides a comprehensive overview of performance metrics for various deep learning models under both "Trainable: True" and "Trainable: False" configurations using holdout validation. When trainable parameters were enabled, models such as DenseNet121, EfficientNetB0, InceptionResNetV2, InceptionV3, MobileNet, and ResNet50 consistently achieved exceptional classification accuracies exceeding 97\%, with EfficientNetB0 reaching over 99\%. This high performance, particularly with fine-tuning, highlights the models' strong adaptability to the transformed Phase-OTDR data. In contrast, freezing the trainable parameters ("Trainable: False") resulted in a noticeable decline in performance across most models. Notably, InceptionResNetV2 experienced a substantial reduction in accuracy, while InceptionV3, MobileNet, and VGG19 also showed significant degradation. This performance disparity underscores that fine-tuning is essential for optimizing model performance on this specific dataset, allowing the models to learn and adapt features relevant to Phase-OTDR event detection rather than relying solely on generic pre-trained features. The ability to achieve high accuracy is critical for practical applications, where precise and reliable event detection directly impacts system effectiveness and operational safety.

\begin{table}[!ht]
\centering
\caption{Performance metrics comparisons for different models under varying trainable parameters using holdout validation.}
\label{tab:t2}
\resizebox{\textwidth}{!}{%
\begin{tabular}{c|cccc|cccc}
\hline
                                      & \multicolumn{4}{c|}{\textbf{Trainable : "True"}}                                                                              & \multicolumn{4}{c}{\textbf{Trainable :   "False"}}                                                                            \\ \cline{2-9} 
\multirow{-2}{*}{\textbf{Model}}      & \textit{\textbf{Cls Acc}}     & \textit{\textbf{Sens}}        & \textit{\textbf{Prec}}        & \textit{\textbf{F1}}          & \textit{\textbf{Cls Acc}}     & \textit{\textbf{Sens}}        & \textit{\textbf{Prec}}        & \textit{\textbf{F1}}          \\ \hline
DenseNet121                           & 0.9824                        & 0.9826                        & 0.9820                        & 0.9821                        & 0.9435                        & 0.9418                        & 0.9421                        & 0.9416                        \\
{\color[HTML]{595959} EfficientNetB0} & {\color[HTML]{595959} 0.9884} & {\color[HTML]{595959} 0.9885} & {\color[HTML]{595959} 0.9880} & {\color[HTML]{595959} 0.9882} & {\color[HTML]{595959} 0.9542} & {\color[HTML]{595959} 0.9535} & {\color[HTML]{595959} 0.9537} & {\color[HTML]{595959} 0.9536} \\
InceptionResNetV2                     & 0.9833                        & 0.9835                        & 0.9831                        & 0.9830                        & 0.4788                        & 0.4925                        & 0.5432                        & 0.4709                        \\
{\color[HTML]{595959} InceptionV3}    & {\color[HTML]{595959} 0.9790} & {\color[HTML]{595959} 0.9785} & {\color[HTML]{595959} 0.9788} & {\color[HTML]{595959} 0.9786} & {\color[HTML]{595959} 0.8904} & {\color[HTML]{595959} 0.8890} & {\color[HTML]{595959} 0.8927} & {\color[HTML]{595959} 0.8898} \\
MobileNet                             & 0.9842                        & 0.9841                        & 0.9836                        & 0.9837                        & 0.8664                        & 0.8633                        & 0.8665                        & 0.8634                        \\
{\color[HTML]{595959} ResNet50}       & {\color[HTML]{595959} 0.9709} & {\color[HTML]{595959} 0.9711} & {\color[HTML]{595959} 0.9700} & {\color[HTML]{595959} 0.9705} & {\color[HTML]{595959} 0.9666} & {\color[HTML]{595959} 0.9665} & {\color[HTML]{595959} 0.9660} & {\color[HTML]{595959} 0.9661} \\
VGG16                                 & 0.9722                        & 0.9718                        & 0.9713                        & 0.9715                        & 0.9456                        & 0.9440                        & 0.9456                        & 0.9442                        \\
{\color[HTML]{595959} VGG19}          & {\color[HTML]{595959} 0.9533} & {\color[HTML]{595959} 0.9529} & {\color[HTML]{595959} 0.9521} & {\color[HTML]{595959} 0.9521} & {\color[HTML]{595959} 0.9289} & {\color[HTML]{595959} 0.9287} & {\color[HTML]{595959} 0.9283} & {\color[HTML]{595959} 0.9281} \\ \hline
\end{tabular}%
}
\end{table}
\FloatBarrier

\begin{figure}[!ht]
    \centering
    \includegraphics[width = 0.95\linewidth]{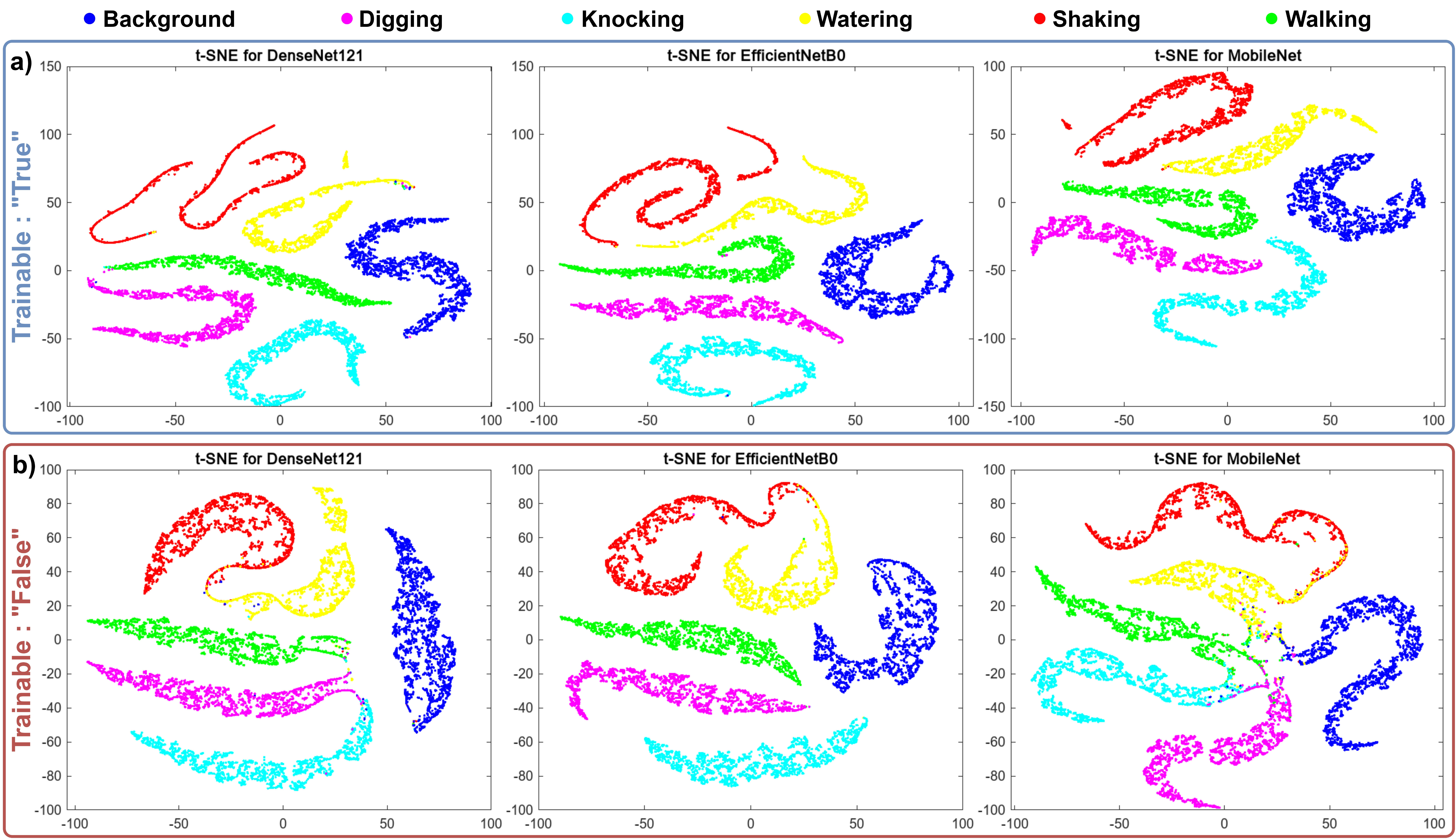}
    \caption{t-SNE visualizations demonstrate the feature space separation for DenseNet121, EfficientNetB0, and MobileNet models with different trainable parameters. The plots represent the clustering of training data as interpreted by each trained model, with the x and y axes depicting the distribution of data points in a two-dimensional space that approximates their high-dimensional proximity.}
    \label{fig:t-SNE}
\end{figure}

The t-SNE visualizations in Figure \ref{fig:t-SNE} illustrate the feature space separation generated by DenseNet121, EfficientNetB0, and MobileNet for six distinct event classes: background, digging, knocking, watering, shaking, and walking. These visualizations qualitatively confirm the quantitative results by showing clear clustering of different event types. Specifically, the "Trainable: True" configuration consistently yields more distinct and well-separated clusters compared to the "Trainable: False" configuration. This improved separation in the feature space directly correlates with the enhanced classification performance observed in fine-tuning mode, indicating that the models learn more discriminative features when their layers are allowed to adapt. The ability to visually distinguish between event classes reinforces the effectiveness of our image-based approach and the fine-tuning strategy in capturing the underlying patterns of Phase-OTDR events.

\FloatBarrier

\begin{figure}[!ht]
    \centering
    \includegraphics[width = 0.8\linewidth]{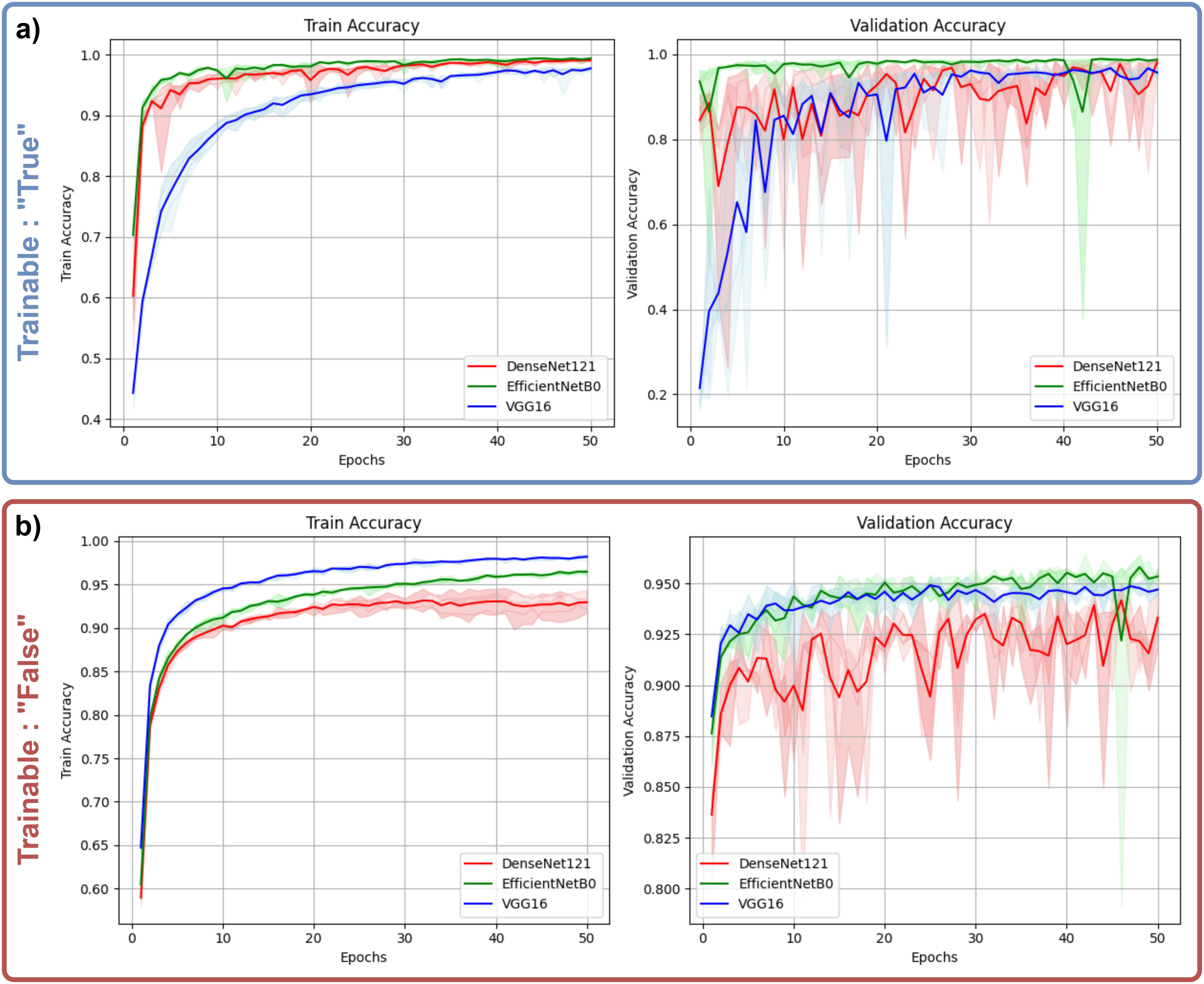}
    \caption{Illustrates the 5-fold cross-validation analysis depicting training and validation curves for selected three distinct models under different trainable parameters. The colored lines represent mean performance metrics across folds, while shaded areas highlight the variations.}
    \label{fig:curve5f}
\end{figure}

\begin{figure}[!ht]
    \centering
    \includegraphics[width = 0.95\linewidth]{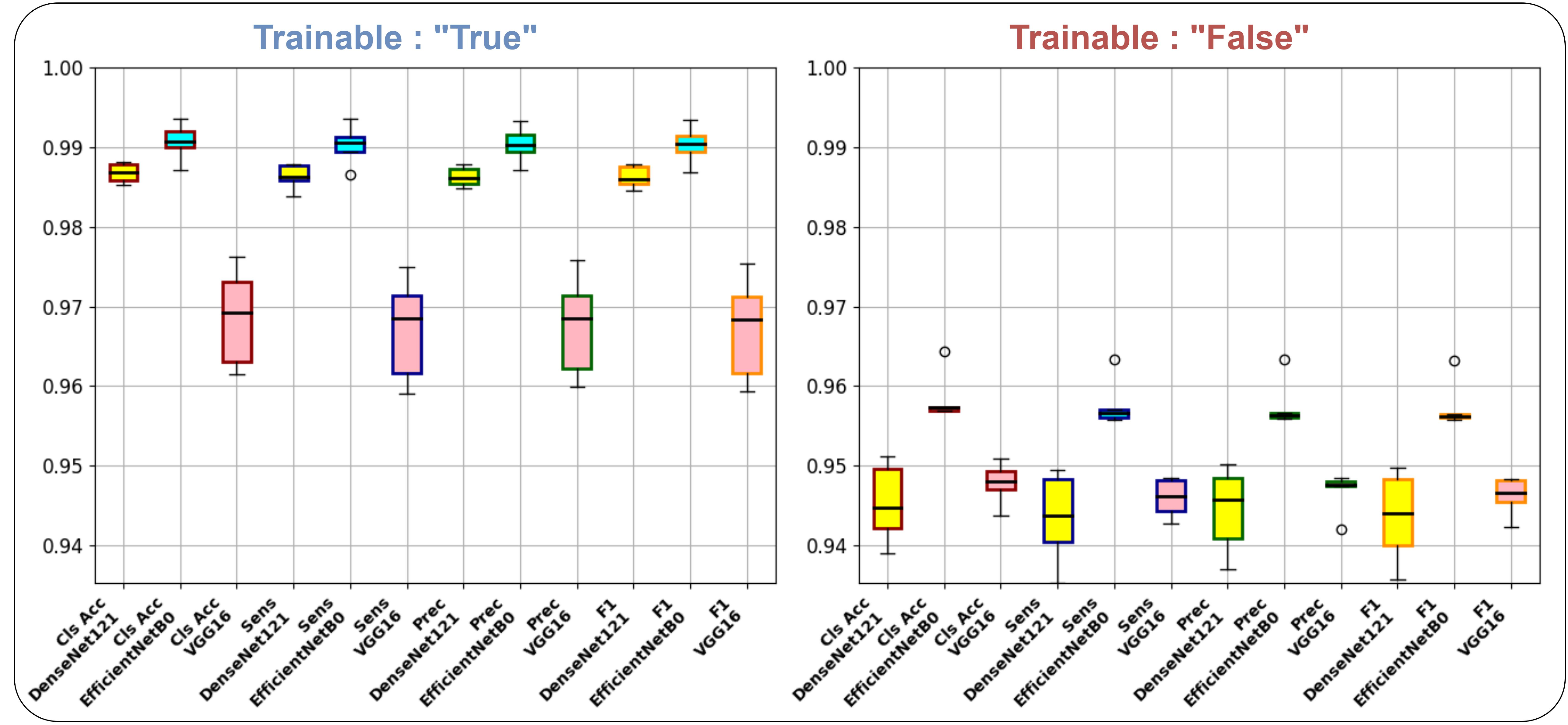}
    \caption{Comparative boxplots depict the performance of transfer learning models with 'Trainable: True' (left) and 'Trainable: False' (right) settings over 5-fold cross-validation. These visualizations compare key performance metrics, including accuracy (Acc), sensitivity (Sens), precision (Prec), and F1 scores, demonstrating the variability and central tendencies in each configuration.}
    \label{fig:box5f}
\end{figure}

The 5-fold cross-validation analysis, depicted in Figure \ref{fig:curve5f} and detailed in Table \ref{tab:t3}, provides a robust evaluation of model stability and reliability. Figure \ref{fig:curve5f} illustrates the training and validation curves for DenseNet121, EfficientNetB0, and VGG16, highlighting consistent performance across different folds. The solid lines represent the mean performance metrics, while shaded areas indicate variations. Table \ref{tab:t3} shows that in the "Trainable: True" configuration, DenseNet121, EfficientNetB0, and VGG16 achieved mean classification accuracies of approximately 98.68\%, 99.07\%, and 96.86\%, respectively. These high mean accuracies, coupled with relatively low standard deviation values (e.g., 0.0012 for DenseNet121 accuracy), demonstrate the models' exceptional stability and generalizability across different data subsets. This consistency across folds implies that the models are not overfitting to specific data partitions and can reliably perform on unseen data, which is a crucial aspect for real-world deployment of Phase-OTDR event detection systems. Conversely, the "Trainable: False" configuration resulted in slightly lower mean accuracies and marginally higher standard deviations, further emphasizing the benefits of fine-tuning for robust and consistent performance. Figure \ref{fig:box5f} further supports these findings through comparative boxplots, visually representing the variability and central tendencies of key performance metrics across both configurations.

\FloatBarrier

\begin{table}[!ht]
\centering
\caption{Performance metrics comparison for three selected models under various trainable parameters using 5-fold cross-validation.}
\label{tab:t3}
\resizebox{\textwidth}{!}{%
\begin{tabular}{c|c|cccc|cccc}
\hline
                                 &                                             & \multicolumn{4}{c|}{\textbf{Trainable : "True"}}                                                                              & \multicolumn{4}{c}{\textbf{Trainable : "False"}}                                                                            \\ \cline{3-10} 
\multirow{-2}{*}{\textbf{Model}} & \multirow{-2}{*}{\textbf{Statistic Type}}   & \textit{\textbf{Cls Acc}}     & \textit{\textbf{Sens}}        & \textit{\textbf{Prec}}        & \textit{\textbf{F1}}          & \textit{\textbf{Cls Acc}}     & \textit{\textbf{Sens}}        & \textit{\textbf{Prec}}        & \textit{\textbf{F1}}          \\ \hline
DenseNet121                      & Mean Acc                                    & 0.9868                        & 0.9863                        & 0.9863                        & 0.9862                        & 0.9453                        & 0.9434                        & 0.9444                        & 0.9435                        \\
                                 & Standard Deviation                         & 0.0012                        & 0.0016                        & 0.0012                        & 0.0014                        & 0.0051                        & 0.0058                        & 0.0054                        & 0.0058                        \\ \hline
EfficientNetB0                   & Mean Acc                                    & 0.9907                        & 0.9903                        & 0.9903                        & 0.9903                        & 0.9587                        & 0.9579                        & 0.9578                        & 0.9577                        \\
                                 & Standard Deviation                         & 0.0024                        & 0.0026                        & 0.0023                        & 0.0025                        & 0.0032                        & 0.0031                        & 0.0031                        & 0.0031                        \\ \hline
VGG16                            & Mean Acc                                    & 0.9686                        & 0.9671                        & 0.9675                        & 0.9672                        & 0.9478                        & 0.9460                        & 0.9467                        & 0.9462                        \\
                                 & Standard Deviation                         & 0.0063                        & 0.0066                        & 0.0066                        & 0.0066                        & 0.0027                        & 0.0025                        & 0.0026                        & 0.0024                        \\ \hline
\end{tabular}%
}
\end{table}

\FloatBarrier

Our study introduces a novel image-based methodology for Phase-OTDR data analysis for event detection, leveraging multi-channel RGB images generated through GADF, GASF, and RP transformations. The fine-tuning approach consistently yielded superior results, with EfficientNetB0 achieving approximately 98.8\% accuracy in holdout validation and 99.1\% in 5-fold cross-validation, and DenseNet121 achieving 98.2\% and 98.7\% respectively. These results are highly competitive when compared to existing deep learning models specifically for Phase-OTDR event detection, as summarized in Table \ref{tab:dl_models_full}. This comprehensive table details various preprocessing methods, model architectures, dataset configurations, and reported accuracies from the literature. Our approach, represented by "Yeke et al. (2025)" in the table, demonstrates comparable or superior performance to many state-of-the-art methods, particularly considering the comprehensive image transformation framework and the use of diverse transfer learning models. For instance, our holdout accuracy of 98.84\% and 5-fold CV accuracy of 99.07\% for EfficientNetB0 are among the highest reported, showcasing the effectiveness of converting 1D Phase-OTDR data into image format for enhanced classification. This underscores the consistency and generalizability of our model, marking a substantial advancement in Phase-OTDR data analysis and holding significant potential across diverse application domains, including critical infrastructure monitoring and security.

\begin{table}[ht]
\caption{Comparative performance of AI models for Phase-OTDR event detection, including preprocessing methods and dataset configurations.}
\label{tab:dl_models_full}
\resizebox{\textwidth}{!}{%
\begin{tabular}{l|l|l|c|c|c|c}
\hline
\multicolumn{1}{c|}{\textbf{Study}} &
  \multicolumn{1}{c|}{\textbf{Preprocessing Method}} &
  \multicolumn{1}{c|}{\textbf{DL Model}} &
  \multicolumn{1}{c|}{\textbf{\begin{tabular}[c]{@{}c@{}}Num. of\\ Classes\end{tabular}}} &
  \multicolumn{1}{c|}{\textbf{Dataset Size}} &
  \multicolumn{1}{c|}{\textbf{Train/Test/Val}} &
  \multicolumn{1}{c}{\textbf{Accuracy (\%)}} \\ \hline
\makecell[l]{Cao et al. \\ (2023) \cite{CAO2023100372}} &
  \makecell[l]{Spatio-temporal matrix normalization, \\ feature extraction (32 features)} &
  SVM, CNN &
  6 &
  15,612 &
  8:2 &
  94.0 \\ \hline
\makecell[l]{Chen et al. \\ (2023) \cite{electronics12183757}} &
  \makecell[l]{Normalization, differentiation, \\ spatio-temporal feature extraction} &
  \makecell[l]{Dendrite Net (DD) + \\ VTTCCG} &
  6 &
  15,612 &
  8:2 &
  98.6 \\ \hline
\makecell[l]{Gan et al. \\ (2024) \cite{Gan22024}} &
  \makecell[l]{Differential processing, \\ Mel-spectrum transformation} &
  \makecell[l]{VGGish + SVM \\ (Transfer Learning)} &
  6 &
  600 &
  8:2 &
  84.17 \\ \hline
\makecell[l]{Hu et al. \\ (2024) \cite{HU2024130818}} &
  \makecell[l]{Normalization (0--255), TST, \\ StyleGAN, Cutout} &
  \makecell[l]{AlexNet \\ (TSC Framework)} &
  6 &
  30,000 &
  5-Fold CV &
  91.0 \\ \hline
\makecell[l]{Kamanga et al. \\ (2024) \cite{KAMANGA2024104032}} &
  \makecell[l]{MFCC-DP(Differential Phase), \\ active learning} &
  \makecell[l]{AlexNet \\ (Modified CNN)} &
  6 &
  13,788 &
  6-Fold CV &
  98.2 \\ \hline
\makecell[l]{Li et al. \\ (2024) \cite{Li2024Aaa}} &
  \makecell[l]{Spatio-temporal feature extraction, \\ channel and spatial attention mechanism} &
  \makecell[l]{MT-ACNN-SA-BiLSTM \\ (Semi-Supervised)} &
  6 &
  15,418 &
  8:2 &
  96.9 \\ \hline
\makecell[l]{Sasi et al. \\ (2024) \cite{10627811}} &
  \makecell[l]{Feature selection, \\ hyperparameter optimization} &
  LightGBM &
  6 &
  15,612 &
  8:2 &
  95.0 \\ \hline
\makecell[l]{Wang et al. \\ (2024) \cite{10557760}} &
  \makecell[l]{Min-Max normalization, \\ FEA-TFW-LDA} &
  \makecell[l]{SVM, RF, KNN, NB} &
  6 &
  15,612 &
  8:2 &
  98.0 \\ \hline
\makecell[l]{Duan et al. \\ (2025) \cite{DUAN2025104171}} &
  \makecell[l]{1D CNN, Modified Transformer \\ Encoder (ISAT), Dropout, Batch Norm} &
  \makecell[l]{ISAT (Inter-sequence- \\ attention Transformer)} &
  6 &
  4,200 &
  $\sim$3:7 &
  98.8 \\ \hline
\makecell[l]{Hu et al. \\ (2025) \cite{HU2025131393}} &
  \makecell[l]{Gaussian noise addition, data \\ reshaping, spatiotemporal maps} &
  SDENet &
  6 &
  15,419 &
  8:2 &
  97.7 \\ \hline
\makecell[l]{Li et al. \\ (2025) \cite{10877714}} &
  \makecell[l]{Long sequence classification, \\ 10,000$\times$12 $\rightarrow$ sequence format} &
  \makecell[l]{DASFormer \\ (Transformer-based)} &
  6 &
  15,418 &
  8:2 &
  99.6 \\ \hline
\makecell[l]{Luo et al. \\ (2025) \cite{10876181}} &
  \makecell[l]{Downsampling, normalization, \\ KD (knowledge distillation)} &
  \makecell[l]{4-layer CNN (student), \\ ResNet-34 (teacher), with KD} &
  6 &
  15,612 &
  8:2 &
  97.92 \\ \hline
\makecell[l]{Wang et al. \\ (2025) \cite{11128983}} &
  \makecell[l]{Mel-spectrogram, normalization, \\ wavelet demodulation} &
  \makecell[l]{IRMS-CNN + Contrastive \\ Prototype Learning} &
  5+1 &
  13,409 &
  6:2:2 &
  \makecell[c]{98.18 (closed), \\ 88.33 (open)} \\ \hline
\makecell[l]{Cheng et al. \\ (2025) \cite{Chengaa2025}} &
  \makecell[l]{TF transformation, masking, \\ pre-training, global-local feature fusion} &
  \makecell[l]{$\Phi$-GLMAE (Global-Local \\ Masked Autoencoder)} &
  6 &
  15,612 &
  -- &
  \makecell[c]{99.74 (pre-train), \\ 98.97 (w/o)} \\ \hline
\makecell[l]{\textbf{Our Study}} &
  \makecell[l]{GADF, GASF, RP $\rightarrow$ RGB \\ Image Transformation} &
  \makecell[l]{8 Different Transfer \\ Learning Models} &
  6 &
  15,612 &
  \makecell[c]{7:1.5:1.5 and \\ 5-Fold CV} &
  \makecell[c]{98.84 (Holdout), \\ 99.07 (5-fold)} \\ \hline
\end{tabular}%
}
\end{table}
\FloatBarrier

\section{Conclusion}\label{sec:4}

This study successfully introduced a novel image-based classification approach that significantly advances Phase-OTDR event detection. Unlike traditional 1D signal processing or basic deep learning architectures, our methodology uniquely transforms 1D temporal traces into multi-channel RGB images using Gramian Angular Difference Field (GADF), Gramian Angular Summation Field (GASF), and Recurrence Plot (RP) techniques. This innovative transformation not only enhances the visual interpretability of complex Phase-OTDR data but also drastically improves data management efficiency by reducing the dataset size from 2.03 GB to 180 MB. Leveraging transfer learning with fine-tuning, our EfficientNetB0 and DenseNet121 models achieved outstanding classification accuracies of approximately 98.8\% and 98.2\% in holdout validation, and 99.1\% and 98.7\% in 5-fold cross-validation, respectively. This superior performance, particularly when compared to models used as mere feature extractors, underscores the critical role of adapting pre-trained networks to the nuanced characteristics of Phase-OTDR event images, thereby enabling more precise and reliable event classification. While this image-based approach marks a substantial advancement, a key limitation is its current reliance on a single, controlled dataset. Future research will focus on validating the generalizability of this methodology across diverse and larger Phase-OTDR datasets, exploring real-time processing implementations, and optimizing image transformation parameters for enhanced robustness in varied operational environments. This research, therefore, presents a compelling argument for the transformative impact of image-based analysis on Phase-OTDR data, paving the way for more accurate and efficient fiber optic monitoring systems in critical infrastructure and security applications.

%
%

\section*{Acknowledgments}
The authors would like to express their gratitude to the \textit{Machine Intelligence Research and Applications Laboratory} and the \textit{Optical Fiber Metrology and Sensors Laboratory} at Izmir Institute of Technology for their valuable support throughout this study.

\section*{Funding}
No external funding was received for this study.

\section*{Author Contributions}
\begin{itemize}
    \item \textbf{M.C. Yeke}: Conceptualization, Methodology, Software, Validation, Formal Analysis, Investigation, Data Curation, Writing - Original Draft, Visualization.
    \item \textbf{Samil Sirin}: Supervision, Resources, Writing - Review \& Editing.
    \item \textbf{Kivilcim Yuksel}: Supervision, Resources, Writing - Review \& Editing.
    \item \textbf{Abdurrahman Gumus}: Supervision, Resources, Writing - Review \& Editing.
\end{itemize}

\section*{Data Availability}
The dataset used in this study includes samples publicly shared by Cao et al..\cite{CAO2023100372} The codes of the study and the corresponding image-based dataset are shared on GitHub in a way that benefits everyone interested: \url{https://github.com/miralab-ai/Phase-OTDR-event-detection}.

\bibliographystyle{unsrt}  
\bibliography{references}

\end{document}